**RESEARCH ARTICLE**

# ReViSe: Remote Vital Signs Measurement Using Smartphone Camera

**DONGHAO QIAO**[1], **AMTUL HAQ AYESHA**[1], **FARHANA ZULKERNINE**[1], (Member, IEEE), **NAUMAN JAFFAR**[2], AND **RAIHAN MASROOR**[3]

[1] School of Computing, Queen's University, Kingston, ON K7L 3N6, Canada
[2] Veyetals, SenSights.AI, MarkiTech, Toronto, ON M5J 2L2, Canada
[3] Your Doctors Online, Toronto, ON L4W 0C2, Canada

Corresponding author: Donghao Qiao (d.qiao@queensu.ca)

The research was supported by grants from Mitacs and Markitech under grant numbers IT26179-FR68762-ON-ISED and IT18892-FR51544-ON-ISED.

This work involved human subjects in its research. Approval of all ethical and experimental procedures and protocols was granted by Queen's University Health Sciences and Affiliated Teaching Hospitals Research Ethics Board (HSREB).

**ABSTRACT** Remote Photoplethysmography (rPPG) is a fast, effective, inexpensive and convenient method for collecting biometric data as it enables vital signs estimation using face videos. Remote contactless medical service provisioning has proven to be a dire necessity during the COVID-19 pandemic. We propose an end-to-end framework to measure people's vital signs including Heart Rate (HR), Heart Rate Variability (HRV), Oxygen Saturation ($SpO_2$) and Blood Pressure (BP) based on the rPPG methodology from the video of a user's face captured with a smartphone camera. We extract face landmarks with a deep learning-based neural network model in real-time. Multiple face patches also called Regions-of-Interest (RoIs) are extracted by using the predicted face landmarks. Several filters are applied to reduce the noise from the RoIs in the extracted cardiac signals called Blood Volume Pulse (BVP) signal. The measurements of HR, HRV and $SpO_2$ are validated on two public rPPG datasets namely the TokyoTech rPPG and the Pulse Rate Detection (PURE) datasets, on which our models achieved the following Mean Absolute Errors (MAE): a) for HR, 1.73Beats-Per-Minute (bpm) and 3.95bpm respectively; b) for HRV, 18.55ms and 25.03ms respectively, and c) for $SpO_2$, an MAE of 1.64% on the PURE dataset. We validated our end-to-end rPPG framework, ReViSe, in daily living environment, and thereby created the Video-HR dataset. Our HR estimation model achieved an MAE of 2.49bpm on this dataset. Since no publicly available rPPG datasets existed for BP measurement with face videos, we used a dataset with signals from fingertip sensor to train our deep learning-based BP estimation model and also created our own video dataset, Video-BP. On our Video-BP dataset, our BP estimation model achieved an MAE of 6.7mmHg for Systolic Blood Pressure (SBP), and an MAE of 9.6mmHg for Diastolic Blood Pressure (DBP). ReViSe framework has been validated on datasets with videos recorded in daily living environment as opposed to less noisy laboratory environment as reported by most state-of-the-art techniques.

## I. INTRODUCTION

Vital signs like the Heart Rate (HR), Heart Rate Variability (HRV), Oxygen Saturation ($SpO_2$) and Blood Pressure (BP) are important indicators of a person's physiological and emotional well-being. HR indicates the number of times a person's heart beats in a minute. HR fluctuates depending on people's physical activity as well as mental state. It can also be indicative of a person's emotional state and reaction to external stimuli [1], [2]. For instance, while playing video games or watching a movie, the HR can indicate how much a person is enjoying or is engrossed in the activity.

HRV denotes the variation in the time interval between adjacent heartbeats. The variances of HR is subtle in relaxed mental and physical states and millisecond is used as the measurement unit to evaluate the HRV which is difficult to measure with a video camera [3]. Variations in light and motion introduce noise and can cause sudden changes in the cardiac signal, which can affect the measurement accuracy of HRV. $SpO_2$ measurement indicates the proportion of oxygenated hemoglobin in the blood compared to the total amount of hemoglobin [4]. It is widely used for monitoring lung infections in patients. $SpO_2$ measurement is based on the theory that Oxygenated Hemoglobin ($HbO_2$) absorbs light differently at different wavelengths than the non-oxygenated Hemoglobin (Hb). BP is the pressure which blood exerts on the walls of the arteries due to the pumping of the heart, which helps circulate the blood through the arteries.

### A. TRADITIONAL METHODS

The most common methods used for monitoring vitals such as Electrocardiogram (ECG), sphygmomanometer, and pulse oximeter require skin contact [5], [6]. ECG, the current gold standard for measuring HR, is not only tedious but can also cause discomfort in a patient as it involves applying gel on a patient's chest in order to measure the heart's electrical signals [5]. Oximeter uses the principle of Photoplethysmography (PPG) wherein the skin is illuminated with light. Proportional to the volume of blood flowing through the tissues, a part of the light is absorbed by the tissues and the rest is reflected. By monitoring the amount of reflected light, the Blood Volume Pulse (BVP) signal is extracted, from which the HR, $SpO_2$, and BP are computed [1], [2]. Although using an oximeter usually does not cause discomfort in the patient, it is not readily available to all for a quick measurement. Mercury and digital sphygmomanometer are widely used for measuring BP. These devices contain an inflatable cuff which is worn on the arm and the BP is measured by increasing and releasing the pressure in the cuff. The pressure applied on the arm causes uneasiness and some people even complain of pain. The most accurate method of measuring BP is to place an arterial catheter in an artery and measure the pressure on a transducer. This method is intrusive and is used only in ICUs for critical patients. It is not usable in daily measurements as it is a very sensitive method which can lead to bleeding and infection if not managed well [37]. Therefore, the medical community is constantly looking for more convenient methods for regular measurement of vital signs to improve health monitoring.

### B. RPPG METHODS

In recent years, the remote methods for measuring the vital signs based on the principle of PPG have gained momentum which proved to be invaluable during the pandemic period. These methods are referred to as rPPG methods [6] and they employ a contactless video based method for vital signs measurement [6]. Computer vision techniques are applied to videos of skin to obtain the BVP signal which represents the blood flow pattern inside the blood vessels [1], [15], [39]. By using statistical models [43], [45] and machine learning algorithms [40], [42], [44] various vital signs can be computed from the BVP signal. The signal obtained by the remote method is very sensitive to motion and illumination artifacts. To eliminate this noise and obtain a stable signal, robust image filtering and signal processing techniques must be applied. rPPG methods can be broadly classified into two methods: the motion-based methods and the intensity-based methods. In the motion-based method, the head movement of the subject is tracked in videos in order to obtain the BVP signal [28], [29]. In the intensity-based method, the light reflected from the skin surface results in slight color changes on the face, which can be tracked in face videos to obtain the BVP signal [9], [10], [11], [12], [13], [14]. rPPG methods are very promising as they not only eliminate the discomfort experienced in intrusive methods, but also provide an ideal solution for enabling remote facial expression analysis and medical consultation.

The major deterrents in obtaining an accurate BVP signal in rPPG are illumination changes in the environment and various motion noises, which include rotation of the head, blinking of the eyes and twitching of the face [1], [9], [45]. These movements are usually present under realistic conditions and are therefore unavoidable. To combat these issues, researchers must employ different techniques in order to filter out noises from the BVP signal due to light intensity changes and motion in the rPPG videos [1], [2], [6].

### C. CONTRIBUTIONS

The key contributions of this work are as follows.

1) We propose an rPPG based end-to-end framework ReViSe as shown in Fig. 1 for remote measurement of multiple vital signs using only a face video captured in the daily living environment using a smart phone camera while the state-of-the-art research uses videos captured in an ideal laboratory environment. The ReViSe framework has been commercialized with further improvements (mobile application is called Veyetals[1]) by our industry partner. Anyone who has a smartphone camera and an access to telecommunication signal can use the mobile ubiquitous service to measure vital signs. The results are returned based on a 20 seconds face video which is extremely useful for remote medical advising.

2) ReViSe measures HR, HRV, $SpO_2$, and BP, all four vital signs unlike the State-of-the-Art (SOTA) approaches which measure mostly HR [1], [2], [9], [10], [11], [12].

3) We develop and present an advanced video data processing pipeline that includes multiple techniques to extract a robust BVP signal such as face detection, face landmarks prediction, extraction of face regions providing high quality signals, intensity-based signal

---

[1] https://veyetals.com

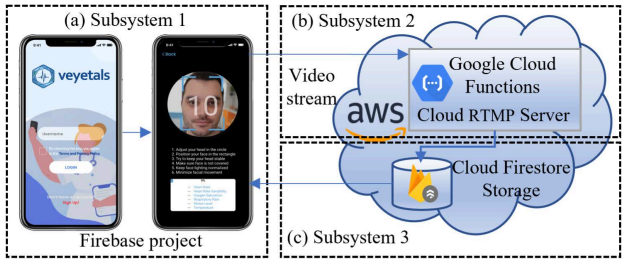

**FIGURE 1.** Overview of the ReViSe framework. It comprises three subsystems: (a) front-end app called Veyetals, (b) back-end processing system and (c) cloud database system.

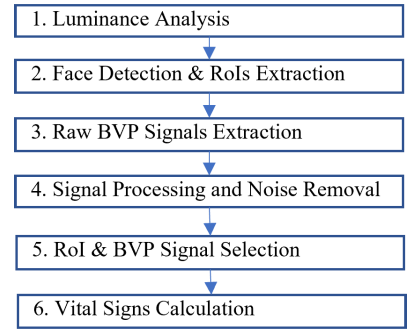

**FIGURE 2.** The six steps performed by the back-end system to process the video data and calculate the vital signs.

extraction, and noise removal techniques to diminish motion and light noises.

This paper presents an extension to our previous work [35], [36] with enhancements in noise removal techniques, the inclusion of an advanced data ingestion and signal processing pipeline as shown in Fig. 1, the addition of BP measurement capability, and the creation of two new datasets, Video-HR and Video-BP. The front camera of a smartphone captures the face video which is streamed to a back-end cloud platform. A deep learning-based face landmarks prediction model processes the streaming video frames and calculates the vital signs. The models are trained and validated using two publicly available datasets namely the TokyoTech rPPG dataset [7] and the Pulse Rate Detection (PURE) dataset [8]. The BP estimation model is trained using a publicly available PPG dataset, and validated using a self-created dataset, Video-BP. The ReViSe framework is validated by volunteers who used a smartphone mobile application to measure their vital signs and simultaneously recorded their vitals using a medical device as the ground truth values.

The rest of the paper is organized as follows. Section II presents an overview of our end-to-end rPPG framework. The related work is discussed in Section III. Section IV explains the methodology in detail. The experiments and results are presented in Section V and Section VI respectively. Finally, Section VII concludes the paper with a brief description of the future work.

## II. OVERVIEW OF THE END-TO-END FRAMEWORK

The end-to-end ReViSe framework is shown in Fig. 1 which facilitates non-invasive remote monitoring of user's vital signs specifically HR, HRV, SpO$_2$, and BP. It is composed of three main subsystems. Subsystem 1 is a mobile application called Veyetals, which captures the user's face video and streams it to the cloud server using the mobile phone service or the internet service. Subsystem 2 is the cloud server at the back-end, which processes the video frames and estimates the vitals. The back-end is linked to a database, which represents Subsystem 3. It stores users data and vitals history. A detailed description of each subsystem is presented below.

### A. SUBSYSTEM 1: FRONT-END APPLICATION - VEYETALS

Veyetals is the front-end mobile application as shown in Fig.1. (a) that is installed on users' smartphones. After the user logs in, the application switches on the front camera and starts streaming the user's video to the cloud platform. When the estimated vital signs are returned by the back end, the application displays the readings to the user on the front-end User Interface (UI), which allows users to interact with the application and view the past measurements.

### B. SUBSYSTEM 2: BACK-END DATA PROCESSING

The back-end hosts a Python program on the Google Cloud Platform (GCP),[2] which accesses the streamed video data at a defined server location through a Universal Resource Locator (URL) as set up in the Amazon Web Services (AWS).[3] Processing of the video data is performed in 6 steps which is illustrated in Fig. 2. In Step 1, we analyze the luminance and brightness of the input video and only the videos with appropriate lighting are processed further. In Step 2, a deep learning-based model is used to detect the face and predict the face landmarks in the video by segmenting it into specific areas referred to as the Regions-of-Interest (RoIs). The changes in the color intensity of pixels in the RoIs yield the raw Blood Volume Pulse (BVP) signals. Multiple RoIs are extracted in Step 2 as the strength of the BVP signals varies in the different face areas such as the forehead, cheeks, and nose owing to varying light conditions, obstruction due to facial hair, and face movement. In Step 3, the raw BVP signals are extracted from all the segmented RoIs. The commonly used RoIs include forehead, cheek, and the area around the nose [7], [8], [13], [14]. In Step 4, the noise due to changes in light intensity and motion is eliminated to a large extent by applying various signal processing and image filtering techniques. This results in robust BVP signals. In Step 5, the RoI that provides the best BVP signal is selected. Finally, the chosen BVP signal is processed further to compute the vital signs namely HR, HRV, SpO$_2$, and BP in Step 6, as explained

---

[2]https://cloud.google.com/
[3]https://aws.amazon.com

**TABLE 1.** A comparative summary of the state-of-the-art rPPG models with different RoI selection and signal processing methods.

| Reference | RoI Detection Methods | Signal Processing | Vital Signs |
|---|---|---|---|
| Rzybyło et al., 2022 [24] | Face detection (whole face) | Long Short Term Memory | HR |
| Song et al., 2021 [25] | Landmarks detection (cheek) | Generative Adversarial Network | HR |
| Tsou et al., 2020 [26] | Face detection (forehead, cheek) | 3D-Convolutional Neural Network | HR |
| Gudi et al., 2019 [14] | Landmarks detection (forehead, cheek) | Motion suppression, bandpass filter | HR, HRV |
| Li et al., 2014 [9] | Landmarks detection (cheek, jaw) | Illumination rectification, non-rigid motion elimination, bandpass filter | HR |
| Wang et al., 2018 [16] | Landmarks detection (face patches) | ICA, bandpass filter | HR |
| Tulyakov et al., 2016 [17] | Landmarks detection (cheek) | Self-Adaptive Matrix Completion | HR |
| Maki et al., 2019 [7] | Landmarks detection (face patches) | ICA, detrending filter, moving average filter | IBI |

later in Section IV. The computed values are transmitted to Subsystem 3 for storage.

### C. SUBSYSTEM 3: CLOUD DATABASE
The computed vital signs of the user are uploaded to a cloud database called Firestore,[4] which is supported by the development platform. The front-end application in Subsystem 1 pulls data from this database and displays the readings to the user. Subsystem 3 allows the vital signs to be saved for registered system users so that the users can monitor the historical measurement data and if needed, share the information with medical professionals for consultation.

## III. RELATED WORK
The proposed framework deploys deep learning models and signal processing techniques to compute the vital signs from users face video. In this section, we describe the relevant literature on face detection and RoI detection, signal processing, and computation of vital signs. A comparative summary of different SOTA rPPG models is shown in Table 1.

### A. FACE DETECTION AND RoI EXTRACTION
The most critical step for accurate measurement of vital signs is selecting an appropriate RoI which can provide a clean and stable periodic BVP signal containing the maximum pulsatile data. The BVP signal is often dampened by motion artifacts owing to involuntary facial movements like blinking, twitching, smiling, and frowning. Therefore, it becomes necessary to choose a RoI that includes the least noise and the most cardiac information.

Two methods are commonly used in the literature for extracting the RoIs on the face [1], [9], [45]. The first method uses a face detector that can localize the bounding box surrounding the face region. The second method suggests an alternative approach to directly predict the coordinates of the face landmarks such as eyes and mouth, which can then aid in segmenting the RoIs. The former approach includes the background along with the face. Since only the facial skin contributes to the cardiac information, it is necessary to eliminate the background. Rectangular RoI was extracted in [8], [10], [11], and [12] to get rid of the background components. 60% width and full height of the detected face is extracted as the RoI. McDuff et al. [8] used a combination of two rectangular face patches and the face landmarks. However, using specific scales of weight and height of the predicted face bounding box to select RoIs can include some background noise. Besides, the face movements or rotations can change the relative regions cropped in the face bounding box.

The face landmarks can help select specific regions on the face and some irregular shaped RoIs were also extracted to exclude parts of the face such as mouth and eyes which are more susceptible to movement. Gudi et al. [14] excluded the mouth, Li et al. [9] excluded the eyes, while Tulyakov et al. [17] excluded both the areas. We attempt to exclude the influence of eyes and mouth, but add the forehead which has a large and flat skin area. In order to diminish the influence of image warping, we use the extracted RoIs directly which is different from [17].

Some researchers selected multiple face regions to overcome incident light intensity and surface reflectance. Maki et al. [7] randomly and repeatedly sampled pair of face patches. Each pair of face patches generated a single raw BVP signal. From these signals, the one with the highest confidence ratio, signal-to-noise ratio and peak height variance was selected. Kumar et al. [15] divided the face into 7 regions by using the face landmarks. The extracted BVP signal was the weighted average of the raw BVP signals from these 7 regions. Wang et al. [16] split the face area into multiple triangular patches and adaptively selected useful patches by analyzing the box plots of pixel quantity. However, the noise and validity of the RoIs are not taken into consideration when selecting the RoIs. Inspired by these methods, we extract three RoIs after face detection including forehead, nose, and face excluding the mouth. This selection approach account for factors such as asymmetric lighting, head rotation and skin occlusion due to facial hair or mask and help select the most effective RoI for vital signs calculation.

In low light condition the skin cardiac data is not clearly visible, which affects the extracted PPG features [9], [10]. Xi et al. [38] decomposed the video frames (L) into an illumination part (T) and a reflectance part (R) which can be expressed as

$$L = R \odot T \qquad (1)$$

---

where $\odot$ means element wise multiplication. Using the green channel as the initial illumination map ($\hat{T}$), they found the enhanced reflectance image (R) by pixel wise division. Guo et al. [53] applied Histogram Equalization (HE) and the low Light Image Enhancement (LIME) via illumination map estimation algorithm to the videos and found that the enhanced videos gave larger RoIs than the original ones and also resulted in improved quality pulsatile signals. The HE-enhanced videos showed very little difference from the LIME-enhanced ones. Quellec et al. [54] improved the image quality by using the YCrCb color space wherein the Y channel is known to represent luminance. Leaving the Cr and Cb components unchanged, the image background (extracted using a Gaussian kernel) was subtracted from the Y component and the resultant image was converted to the RGB color space.

### B. SIGNAL PROCESSING

The signal processing in rPPG is a vital step in generating denoised and clean BVP signals. Signal processing methods can be divided into deep learning-based and conventional ones. The deep learning-based methods use deep learning neural networks to regress a BVP signal from the original video clips or raw BVP signals.

Przybyło [24] and Song et al. [25] extracted a raw BVP signal from the original face video, and leveraged LSTM and GAN to process the raw BVP signal and regress the BVP signal respectively. Tsou et al. [26] utilized 3D-CNN to directly regress the BVP signal from the video which is very time consuming. The input of the network has two branches: video of the forehead and video of the cheek. A shared-weight 6 layer 3D-CNN is applied to regress the two BVPs separately and the sum of the two BVPs is utilized to estimate HR. Deep learning-based methods can achieve comparable measurements to the conventional methods. However, these methods are data intensive and may have significant influence on prediction results depending on the quantity and quality of the training data. Therefore, the conventional signal processing methods for noise reduction and BVP signal generation were explored further in this work.

In conventional rPPG methodology, the individual video frames are monitored over a period of time to track the changes in pixel color intensity or location of facial features. These changes generate the raw BVP signal. Once the RoI is selected, two main techniques are generally applied to extract the raw signal: motion-based methods and color intensity based methods [9]. Efficient signal processing is critical for eliminating noise and accurately calculating the vital signs. While the motion-based methods track coordinates of facial points from frame to frame, the light intensity-based methods track variations in pixel intensity between video frames which indicate the color changes due to cardiac activity.

Balakrishnan et al. [28] proposed the use of head movement for obtaining the BVP signal. The motion-based methods compute the ballistocardiogram movement obtained from the head movement. The involuntary vertical head movement happens due to the pulse and the bobbing movement happens due to respiration. The basic idea is to track features from a person's head, filter out the velocity of interest, and then extract the periodic signal caused by heartbeats. The featured points on the face (face landmarks) represented by coordinates $x(n)$ and $y(n)$, are tracked along time, and only the vertical component $y$ is used to extract the trajectory. The trajectory is used as the raw signal.

However, the intensity-based methods are more popular due to the ease of implementation wherein many techniques can be applied to extract color changes caused by the pulse. When the skin is illuminated by light, the absorption and reflection of light is a function of the blood volume flowing through the capillaries in the region due to heartbeat. The hemoglobin tends to have a higher absorption capacity in the green color range of light than the red color channel [46]. However, all the three-color channels namely red (R), green (G) and blue (B) contain PPG data. Consequently, the green component is widely used to extract the PPG data. Blind Source Separation (BSS) is used if all the three-color channels (RGB) are used.

Wang et al. [9] utilized the PPG data from the green channel of the RGB signal as it has the maximum Signal to Noise Ratio (SNR). Poh et al. [10], [11] used all three-color channels with BSS for signal extraction. Li et al. [9] applied the detrending filter to remove the low frequencies from the signal. To further eliminate the irrelevant frequencies, various bandpass filters were used. Balakrishnan et al. [28] used Butterworth filter while Li et al. [9] used Finite Impulse Response (FIR) Bandpass filters with cutoff frequency representing the normal HR.

BSS is used to separate a mixed signal into its source components [7], [10], [11], [12]. This technique can help separate the noise component from the BVP signal. The two commonly employed BSS techniques are Principal Component Analysis (PCA) [18] and Independent Component Analysis (ICA) [19]. Among the various ICA algorithms, Joint Approximate Diagonalization of Eigen matrices (JADE) is popular for HR calculation as it is computationally efficient [11], [12]. Balakrishnan et al. [28] used PCA for extracting the head trajectories caused by cardiac activities.

ICA involves separating a multivariate signal into its additive subcomponents. The use of signal from the green channel and separating it using ICA is the most popular method employed in rPPG. Assume that a linear mixture of observed signal $x = As$. The ICA approach first finds a matrix $W$, which is the inverse of $A$, and then obtains the independent components by $s = Wx$. PCA maps the signal into unit vectors (known as eigen vectors) where each vector is in the direction of a line that best fits the data while being orthogonal to the other vectors. The eigen vectors are in the directions along which the features have maximum variance. In intensity-based methods, the aim is to extract the component with the highest variance (first eigen vector). In motion-based methods, the component with the highest periodicity in frequency spectra is chosen as the BVP signal [47]. PCA has lower computational complexity compared to ICA, while ICA is more robust [1].

SNR is the ratio of the first two harmonic ranges of the signal to the remaining parts of the signal in the frequency domain. It is used as a metric to analyze the quality of the BVP signal [38], [53]. For motion elimination, Haan et al. [55] divided the signal into $m$ segments of equal length and computed the standard deviation of each segment. The top 5% of the segments with the highest standard deviation were eliminated and the remaining segments were concatenated.

### C. VITAL SIGNS CALCULATION

We calculate four different vital signs from the extracted BVP signal: HR, HRV, SpO$_2$, and BP as described below.

#### 1) HR

The time-domain algorithms [10], [11], [13], [14] detect the peaks in the BVP signal and compute the HR based on the interval of the peaks. However, these methods rely heavily on the extraction of a clean BVP signal and are sensitive to noise. Since the HR is usually a periodic signal, the BVP signal can be transformed into the frequency domain using the Fast Fourier Transform (FFT) [9], [12] or Discrete Cosine Transform (DCT) [29] methods. Machine learning algorithms like Support Vector Machines (SVM) were used by Monkaresi et al. [30] and Osman et al. [31] to compute the Inter-Beat Interval (IBI) from the power spectrum. Deep learning models like DeepPhys [32] and HR-Net [33] extracted BVP signals for HR calculation from a series of images in videos. However, a large dataset is needed to train deep learning models using supervised learning algorithms.

#### 2) HRV

HRV can be computed by calculating the time interval between two successive peaks in the BVP signal in the time-domain. This process is extremely sensitive to noise as random spikes and jumps can add artifacts to the signal and thereby, affect the accuracy of the results. There are also some other HRV measurement techniques applicable to the frequency domain such as Low-Frequency (LF) power, High-Frequency (HF) power, and ratio of LF-to-HF power.

#### 3) SpO$_2$

Tamura et al. [4] described the principle of SpO$_2$ estimation [34]. They used red light and infrared light for calculating SpO$_2$. For SpO$_2$, pixel intensities of two different light sources of different wavelengths are approximated using mathematical models. SpO$_2$ computation is based on the principle that the absorbance of Red (R) light and Infrared Red (IR) light by the pulsatile blood (blood flowing with periodic variations) changes with the degree of oxygenation. The extracted BVP signal obtained from the reflected light is divided into two parts: the Alternating Current (AC) component resulting from the arterial blood and the Direct Current (DC) component resulting from the underlying tissues, venous blood and constant part of arterial blood flow.

The DC component is subtracted while the AC component is amplified of the R and IR lights and used to calculate the SpO$_2$ level in the blood using Eq. 2 as given below.

$$SpO_2 = A - B \times \frac{AC_R/DC_R}{AC_{IR}/DC_{IR}} \qquad (2)$$

where $A$ and $B$ are parameters that can be calibrated by using a standard pulse oximeter.

#### 4) BP

A non-linear relationship exists between the temporal features of a BVP signal and the BP [37]. Therefore, different machine learning models have been employed to exploit this non-linearity. Chowdhury et al. [40] used the PPG signal recorded with a fingertip sensor device to extract 107 features such as the systolic and diastolic peak, notch, systolic peak time, first and second derivative of the signal, demographic features (height, weight, BMI, gender, age), statistical features (standard deviation, skewness, kurtosis), time domain features and frequency domain features. Using ReliefF feature selection [41] for automatic attribute selection and Gaussian Process Regression (GPR), they estimated the SBP and DBP. Su et al. [48] used a sequence learning model based on a deep recurrent neural network for predicting multi day BP from seven handcrafted features of ECG and PPG signals. Autoencoders were used to extract the complex signal features by Shimazaki et al. [50] to train a four layer neural network. Xing et al. [42] and Viejo et al. [51] used the PPG signal's amplitudes and phases as input to a feed forward neural network. Slapnicar et al. [49] trained a random forest model and a deep neural network (Spectro-temporal ResNet) with PPG signals from MIMIC-III dataset. The first model used frequency domain features while the second model used the signal, its first and second derivatives, and the frequency features calculated using spectrogram. They reported that the deep learning model was the best model for BP prediction with an MAE of 6.7 for SBP and 9.6 for DBP. Huang et al. [43] used the results from applying transfer learning on the MIMIC II dataset with k-nearest neighbours for BP prediction from face videos. Schrumpf at el. [52] trained Resnet, AlexNet, Spectro Temporal ResNet and LSTM models. They trained the models on PPG data from MIMIC-III dataset and then used transfer learning to train on rPPG data. They found the Resnet gave the best performance.

## IV. METHODOLOGY

This section first describes the functionality of the end-to-end ReViSe framework and then explains the methods applied in the back-end cloud system to measure the vital signs from the streaming video data.

### A. END-TO-END FRAMEWORK: ReViSe

We developed a smartphone and tablet compatible mobile and cloud supported end-to-end framework in collaboration with our industry partner, which has been productized as the mobile application called Veyetals. The front-end user interface for the mobile device has been developed in React Native[5] for iOS and Android mobile platforms. The complete framework has been developed using the Firebase platform which has a cloud server back-end. The mobile application must be downloaded from app store and deployed on a smartphone or mobile device. As the user starts the application by clicking on the application icon, the front camera from the user's smart device gets activated to capture user's face video for measuring vital signs. The user gets about 10 seconds to read the instructions and adjust the device camera such that the face is positioned within a rectangle box on the screen (Fig. 1. (a)). This front-end application records a 20-second video and simultaneously down-samples it to $1280 \times 720$ px resolution to reduce the volume of data that must be transferred over the communication network. The down-sampled video is recorded and streamed simultaneously at 30 Frames-Per-Second (fps) to a Real-Time Messaging Protocol (RTMP) server at the back-end set up on the cloud using AWS.

The back-end cloud server system constitutes of multiple AI models and algorithms to process the video, compute the vital signs, and return the measurements back to the user's device. Python3 programming language is used to develop the models and algorithms, which are compiled, tested and deployed as the back-end application on the Google Cloud Platform (GCP) as a function. Videos streamed to the RTMP URL are accessed by the back-end application developed in Python using Google Cloud Function when triggered by React Native. Before processing, the brightness of the video is assessed to ensure that a good video with strong rPPG features is used to compute the vital signs as described in Section IV-B.

As soon as the processing completes, the results are stored in the Firebase Firestore database, a real-time cloud storage system in Subsystem 3, which is supported by React Native. The front-end React Native application then reads the measurements from the cloud-based Firestore database and displays the same to the user on the front-end mobile interface.

### B. BACK-END VIDEO PROCESSING AND VITAL SIGNS CALCULATION

The video is processed to calculate vital signs in a 6-step pipeline in the back-end system as shown in Fig. 2.

#### 1) STEP 1: LUMINANCE ANALYSIS

After receiving the video, we analyze the luminance and brightness condition with following methods.

- Mean Grayscale - The RGB image of a video frame is converted to grayscale and the mean of all pixel values is computed. The range of pixels in grayscale is 0-255 and 127 lies in the midway. Therefore, if pixel values lie between 127-255, the video frame corresponds to a light image. However, it was empirically determined that HE improved the brightness resulting in good PPG data when the mean is $< 75$. When the mean is between 75-127 HE added noise to the PPG data.

- Low contrast - Using skimage.exposure library's function is_low_contrast,[6] the video frame contrast is computed and compared with a threshold of 0.65. If the image contrast is low, the lighting is not good enough to return the PPG features.

- Y channel of YCrCb model - The RGB image is converted to YCrCb color space and the mean value of Y channel pixels is computed. This value is monitored in all the video frames and if it varies beyond a threshold of 15 units, the system returns a message that the ambient light is varying.

#### 2) STEP 2: FACE DETECTION AND RoIs EXTRACTION

Considering the robustness and efficiency of face detection and face landmarks prediction, the open source deep learning-based MediaPipe Face Mesh framework [23] is applied to process the user-video frame by frame. First, the Blaze-Face [22] is applied to extract the whole face area in the image. The face detector consists of a Convolutional Neural Network (CNN) with BlazeBlock for feature extraction and a modified Single Shot multibox Detector (SSD) for bounding box prediction.

After detecting the face, a CNN is applied to extract features from the face image and 478 face landmarks are predicted on the face. The face landmarks are used to segment 3 RoIs as shown in Fig. 3: RoI 1 is the forehead, RoI 2 is the cheek and nose area, and RoI 3 is the most face area excludes mouth. A desired RoI is the one that contains the most facial skin as the skin includes the maximum cardiac signal, while having the least noise. RoI 1 is the forehead area which has the largest and flattest skin on our face, it can also be used when users are wearing a mask. However, when user's hair covers the forehead (as shown in Fig. 3), this RoI contains less useful information and more noise as indicated by the larger variance in the extracted signal in Fig. 3. (a). RoI 2 is utilized due to obstruction by hair and/or beard, and motion noise from eyes blinking or mouth movements. RoI 3 aims to select the maximum skin to extract the general information from the face.

#### 3) STEP 3: RAW BVP SIGNAL EXTRACTION

Each RoI provides a three dimensional array comprising of pixel intensities in the three color channels: red, green and blue. The raw BVP signals of each frame are represented by the mean pixel values of the green channel. Hence, the 3 RoIs generate 3 raw BVP signals $g_1(n)$, $g_2(n)$, and $g_3(n)$ where $n = 1, 2, 3, \ldots$ is the index of frame. Next, we process the raw BVP signals to eliminate the noise and extract final clean BVP signals.

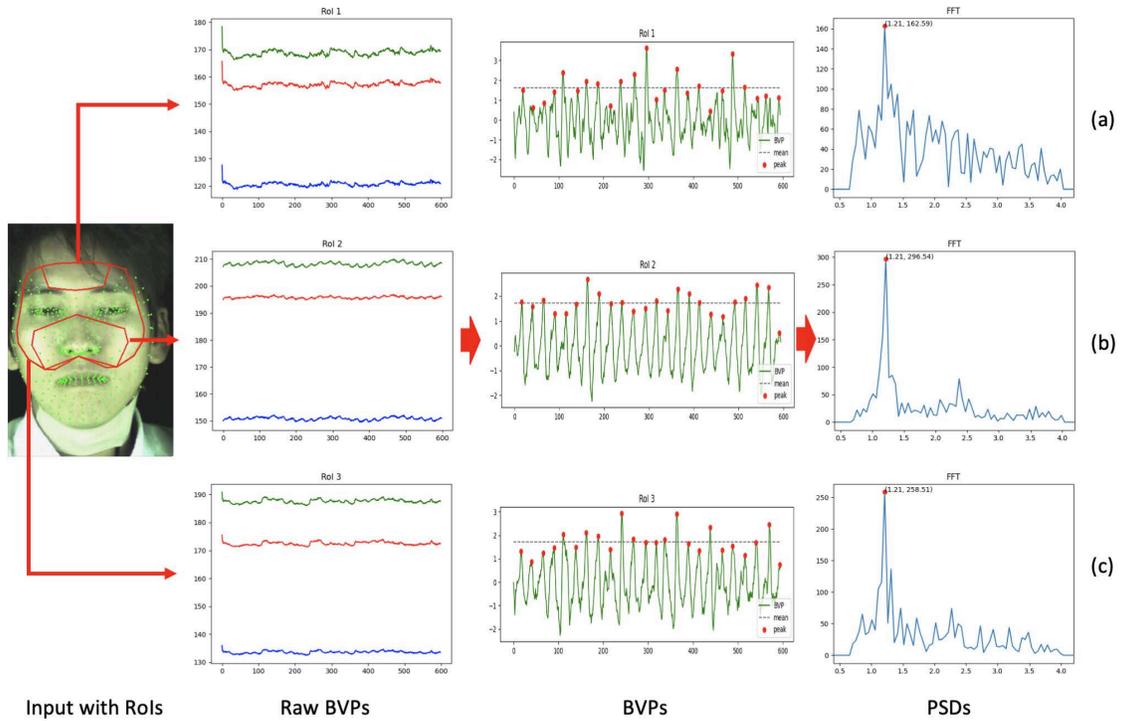

**FIGURE 3.** Video processing pipeline includes the face landmarks prediction, raw BVPs extraction from the segmented RoIs, BVPs extraction with signal processing algorithms and PSDs calculation by using the BVPs. The green dots on the input image (first column) are the predicted face landmarks and the three RoIs are highlighted in red. The second column showing the three raw BVPs from three RoIs include: (a) signals from RoI 1 (forehead); (b) signals form RoI 2 (nose and cheek); (c) signals from RoI 3 (face excluding mouth). The three red, green and blue signal waves in the raw BVPs correspond to the RGB channels. The raw BVP signal of RoI 1 is more variant since the skin is covered by hair. The third column displays the extracted BVPs and corresponding peaks (red dots). The Power Spectral Densities (PSDs) are shown in the fourth column.

### 4) STEP 4: SIGNAL PROCESSING

To attenuate the noise and extract the desired BVP signal, the raw BVP signal is processed in several steps as explained below.

- Signal to Noise Ratio (SNR): To analyze the quality of the signal we use the SNR metric. We transform the raw BVP signal to the frequency domain and only retain the components between 30-240 Hz as this is the human HR range. The components are then normalized and the SNR is calculated using Eq. 3.

$$SNR = 10 \log_{10} \left( \frac{\sum_{f=30}^{240} (U_t(f)\hat{S}(f))^2}{\sum_{f=30}^{240} 1 - (U_t(f)\hat{S}(f))^2} \right) \quad (3)$$

where $\hat{S}(f)$ is the spectrum of the BVP signal S, f is the frequency in bpm, and $U_t(f)$ is a binary template window that extends from the maximum amplitude to 3 units after that. This has been illustrated in Fig.4. Therefore, the signal corresponds to the amplitudes inside the binary window and the remaining amplitudes are the noise.

The SNR is computed for each RoI. When the SNR is negative, a utility to eliminate the motion artefacts is carried out. In here, the time domain signal is divided into 10 segments of equal length. The standard deviation of each segment is calculated and the top 5% segments with the highest standard deviation are eliminated. The resultant signal is the cleaned raw BVP signal.

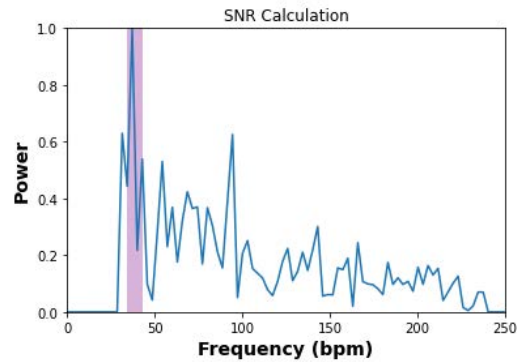

**FIGURE 4.** The SNR is the ratio of the components inside to those outside the binary window of 3 units.

- Denoise Filter: We apply a customized denoise filter to the raw BVP signal to remove the large jumps and steps caused by motion noise such as head rotation and shaking. The input is the time series BVP signal $S$. We calculate the absolute difference of the data points in the signal consecutively and compare this difference against a threshold $v$, where $v = 1$. If the threshold is exceeded, we subtract the difference from the posterior signal to remove the steps. The pseudocode of our denoise filter is shown in Algorithm 1.

---

**Algorithm 1** Denoise Filter

---

**Input:** Signal $S$ and threshold $v$
**Initialize:** integer $i \leftarrow 0$, list $S_{new} \leftarrow [S[0]]$
**while** $i < len(S) - 1$ **do**
    **if** $abs(S[i + 1] - S[i]) > v$ **then**
        $S[i + 1 :] \leftarrow S[i + 1 :] - (S[i + 1] - S[i])$
    **else**
        $S_{new}.append(S[i + 1])$
        $i \leftarrow i + 1$
    **end if**
**end while**
**Output:** $S_{new}$

---

- Normalization: Normalization makes the process of comparison of two signals more robust as it brings the signals within the same range. We normalize the denoised signal by subtracting its mean and then dividing the result by the standard deviation as shown in Eq. 4.

$$g_i(n) = \frac{g_i(n) - \mu_i}{\sigma_i} \quad (4)$$

where $\mu_i$ and $\sigma_i$ are the mean and standard deviation of the signal $g_i(n)$ respectively.

- Independent Component Analysis (ICA): Next, to extract the independent source signal from the mixed signal set, we use ICA. ICA randomly returns a positive or a negative signal, so that the output signal may sometimes be reversed after ICA. This is of little significance in HR calculation in frequency-domain, but to estimate the HRV in time-domain, more accurate peak positions are required. Therefore, we return the signal (positive or negative) that has a higher correlation with the input signal.

- Detrending Filter: After obtaining the source signal, we apply a detrending filter [20] which is designed for PPG signal processing. Detrending filter helps in reducing the non-stationary components of the signal. The method is based on smoothness priors formulation and a smoothing parameter $\lambda = 10$ is applied to adjust the frequency response.

- Moving Average Filter: Finally, a moving average filter as shown in Eq. 5, is applied with $L = 5$ to remove random noise. Moving average filter helps in temporal filtering of the signal. It computes the average of the data points between the frames, thereby reducing random noise yet retaining a sharp step response.

$$g_i(n) = \frac{1}{L} \sum_{k=0}^{L-1} g_i(n - k) \quad (5)$$

where $L$ is the number of points and $n$ is the index of frame. This filter helps in smoothing the signal by removing jumps due to sudden light changes or motion.

### 5) STEP 5: RoI AND BVP SELECTION

Signal processing provides 3 BVP signals $g_1(n)$, $g_2(n)$, and $g_3(n)$ corresponding to the 3 RoIs. The Power Spectral Density (PSD) of each BVP is computed with the Welch's method [21]. The signal with the highest peak spectrum power is selected for calculating the vital signs. Fig. 3 shows the PSDs of the three BVP signals and the peak values of these PSDs, which are 162.59 of RoI 1, 296.54 of RoI 2 and 258.51 of RoI 3 respectively. Therefore, RoI 2 with the highest PSD and its corresponding BVP 2 will be selected to calculate the vital signs.

### 6) STEP 6: VITAL SIGNS CALCULATION

- HR Calculation: HR is normally a periodic signal. So the HR is calculated in the frequency-domain by using the PSD of the selected BVP signal. In Fig. 3. (b), signal 2 is selected as its power is the maximum. The peak coordinate (1.21, 296.54) indicates the HR frequency $f_{HR}$ is 1.21 and the maximum power is 296.54. The final calculated HR of the subject is $60 \times f_{HR}$ bpm i.e., $60 \times 1.21 = 72.6$bpm.

- HRV Calculation: In time-domain, Inter-beat Interval (IBI) is an important cardiac parameter used to calculate HRV. IBI is the time period between the heartbeats represented by the peaks of the extracted BVP signals. The red dots of the BVPs in Fig. 3 are the peaks, which must be greater than zero. Therefore, $IBI = t_n - t_{(n-1)}$ where $t_n$ is the time of the $n$ th detected peak. The Root Mean Square of Successive HR Interval Differences (RMSSD) is calculated using Eq. 6, which is one of the HRV time-domain measures [3] that represents the HRV.

$$RMSSD = \sqrt{\frac{1}{N-1} \sum_{i=1}^{N-1} (IBI_i - IBI_{i+1})^2} \quad (6)$$

where N is the number of IBIs in the sequence.

The HRV measurement is very sensitive and vulnerable in time-domain. It heavily relies on a clean and reliable BVP signal and accurate peak detection. Therefore, after getting a set of IBIs from the BVP signal, we set one standard deviation from the mean of the set as the cut-off in identifying and removing the outliers, and diminish the influence of calculation errors caused by peak detection and residual noise.

- SpO$_2$ Calculation: As we discussed in the related work Section III-C, we utilized the AC and DC of the R and IR signals to calculate the SpO$_2$ with Eq. 2. The calibration parameters $A$ and $B$ are 1 and 0.04 respectively in our experiment.

- BP Calculation: We trained a deep learning model using ResNet blocks for predicting the BP. Since the face video samples were limited, transfer learning was employed.

The model was trained on BVP signals obtained first from PPG samples recorded with a sensor and then from rPPG videos. The network has three branches for the three different input signals namely the BVP signal, its first derivative, and its second derivative. After passing the signals through the ResNet blocks, the outputs were flattened, concatenated and passed through two dense layers to generate the SBP and DBP as output. The network architecture is shown in Fig. 5.

## V. EXPERIMENTS

In this section we present the experiments that we conducted to train and test our framework mainly focusing on the robustness and accuracy of the models for daily living use case environment.

We recruited participants with the help of our industry partner to use the smartphone application for demonstrating the usability and performance of the framework in daily living environment. The data collected was used for testing the models trained with benchmark data. In this paper we mainly present the quantitative evaluation for the initial version of our application where we compared the values returned by our framework against ground truth data measured using a medical device.

Most of the existing work have only validate their models in controlled laboratory environment [1], [9], [57] which generally provides good quality video to start with. We validated our system in two phases, (i) using benchmark dataset, and (ii) using our own dataset created from videos and ground truth data contributed by a group of participants with their informed consents.

### A. EXPERIMENT DESIGN

We designed 4 experiments to evaluate our models for HR, HRV, SpO$_2$, and BP measurement. These models were deployed on the back-end cloud server. The HR, HRV and SpO$_2$ compute models were evaluated using two publicly available rPPG datasets as illustrated in Experiments 1 and 2. In Experiment 3, additional tests were done using a dataset, Video-HR, created from the data which was collected from a group of participants who used the Veyetals smartphone application to measure their vital signs while recording the same using a medical device. The face videos were recorded in daily living environment with ambient lights. Finally, Experiment 4 presents the BP measurement model development and validation results. Experiment 4 was conducted with a public benchmark dataset and the Video-BP dataset that we created.

### B. BENCHMARK DATASETS

Three publicly available datasets were selected to train and test our models. The data were collected in a laboratory environment using stable light sources, so the videos generally have good luminance. The TokyoTech rPPG Dataset [7] contains PPG data recorded using finger-contacted PPG sensors that measured HR and HRV values. The other dataset,

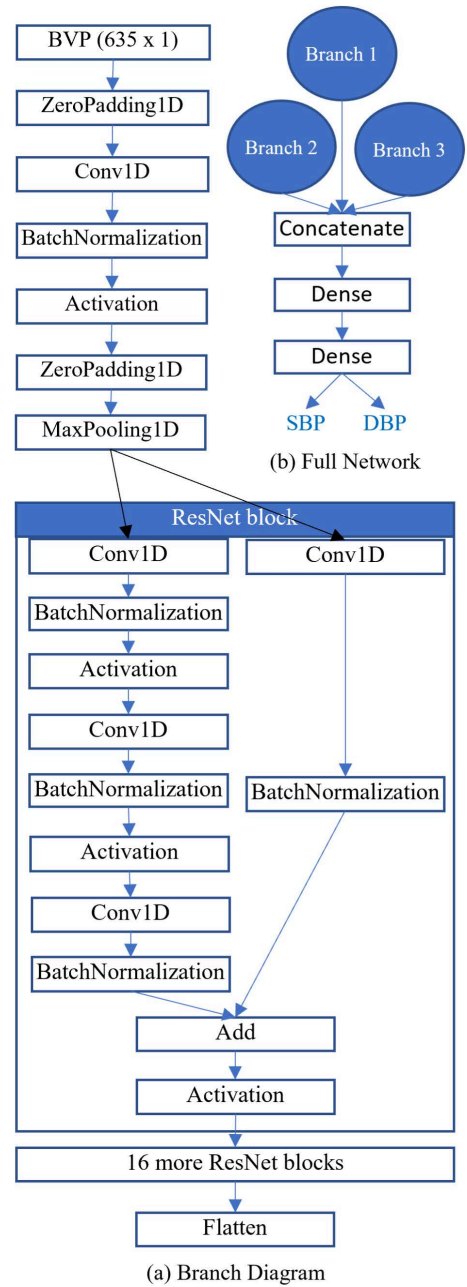

**FIGURE 5.** Deep learning ResNet model architecture for BP Estimation taking three inputs: the BVP signal, its first derivative and its second derivative. a) The branch diagram illustrating the layers through which each input passes, b) the output of the three branches are concatenated and passed through Dense layers to get SBP and DBP.

PURE [8], contains the ground truth values of HR, HRV and SpO$_2$. This dataset is more challenging since some of the data contain face videos with various head movements and rotations. The models can be leveraged to achieve higher accuracy for this data to ensure robustness and reliability against motion noise interference. A third PPG-BP dataset published by Liang et al. [56] was used for training the BP model. A summary of the datasets is shown in Table 2 including the number of participants, number of videos, ground truth vital signs and data collection environment. The detailed description of these datasets is given below.

**TABLE 2.** Summary of rPPG datasets used for evaluation.

| Dataset | Subjects | Videos | Vital Signs | Environment |
|---------|----------|--------|-------------|-------------|
| TokyoTech [7] | 9 | 27 | HR, HRV | Laboratory |
| PURE [8] | 10 | 60 | HR, HRV, SpO2 | Laboratory |
| Video-HR | 15 | 30 | HR | Daily Life |
| Video-BP | 49 | 144 | BP | Daily Life |

**TABLE 3.** Performance on TokyoTech rPPG Dataset [7].

| | Relax | Exercise | Relax | Overall |
|---|-------|----------|-------|---------|
| HR (bpm) | 1.29 | 2.54 | 1.34 | 1.73 |
| HRV (ms) | 16.66 | 20.85 | 18.12 | 18.55 |

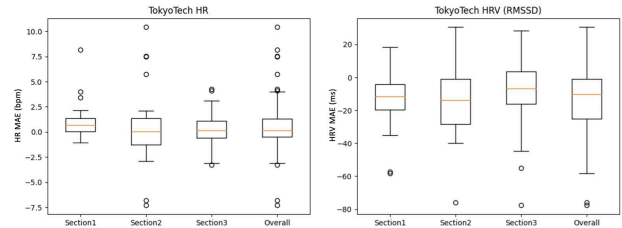

**FIGURE 6.** Boxplot of the MAE in HR and HRV (RMSSD) calculation on the TokyoTech rPPG dataset [7]. The model performs well on the relax sections with less variation, but the exercise increases the MAE in both HR and HRV estimation.

- TokyoTech rPPG Dataset: This dataset consists of 9 subjects (8 male and 1 female) between the age group of 20 to 60 years. Each subject has three 1-minute videos corresponding to three sessions: relax, exercise and relax. The participants perform hand grip exercise before the exercise session. Each 1-minute video is split into three 20-second videos. A finger clip Contact PPG (cPPG) sensor (Procomp Infinity T7500M, Thought Technology Ltd., Canada) with a sampling frequency of 2048 Hz is used to gather the contact BVP signals for ground truth reference.
- PURE Dataset: This dataset contains 10 subjects (8 male and 2 female) and each subject has 6 different setups of 1-minute videos. Therefore, there are 60 video sequences and the total video duration for each subject is 6 minutes. It is more challenging to achieve good results with this dataset as the videos were recorded in 6 setups including steady, talking, slow translation, fast translation, small rotation and medium rotation. The videos were captured with an eco274CVGE camera by SVS-Vistek GmbH at a frame rate of 30 fps with a cropped resolution of 640 × 480 pixels. A finger clip pulse oximeter (pulox CMS50E) is applied to simultaneously measure pulse rate wave and SpO2 with a sampling rate of 60 Hz.
- PPG-BP Dataset: Due to the absence of a dataset containing face video and ground truth BP, we used the publicly available PPG-BP dataset released by Liang et al. [56]. It contains 657 PPG signal samples collected from 219 subjects using finger PPG sensor. The dataset also contains ground truth vitals namely HR, SBP and DBP recorded simultaneously.

### C. SELF-CREATED VIDEO-HR DATASET

To evaluate the end-to-end system, we built our own dataset named Video-HR with the help of 15 participants between 10 to 80 years of age using their smartphones with our framework to measure their vital signs in daily living environment. Each participant recorded two 20 seconds face videos with his/her smartphone Veyetals application in different light and active positions. The smartphones used were Samsung Galaxy Note 9, iPhone 6s, iPhone X and iPhone 11 Pro. Ground truth HR measurement were collected at the same time using pulse oximeter for this pilot study. In the future we plan to conduct the experiment in a clinic where the ground truth data would be collected using authorized personnel using more accurate and rigid medical devices.

### D. SELF-CREATED VIDEO-BP DATASET

To the best of our knowledge there is no open public dataset that contains BP readings with users' face videos. Therefore, we created a dataset named Video-BP for training the BP estimation model. The samples in this dataset include face videos which were recorded by a group of participants using a Samsung Galaxy Note 9 smartphone camera application for about 25 seconds at a frame rate of 30 fps in an environment with ample daylight. The recording device was either held in hand, placed on a surface, or fixed on a tripod stand. This was done to regularize the model to adapt to device movement while video recording. The ground truth BP was recorded using the Andesfit Health BP Monitor. 49 people comprising of 15 males and 34 females between 11-78 years of age participated in the study which resulted in 144 data samples. In the collected dataset, the SBP ranges between 85-168mmHg and the DBP ranges between 50-103mmHg.

### E. MODEL TRAINING AND VALIDATION

The metric used to evaluate the performance of the vital signs measurement models is Mean Absolute Errors (MAE).

#### 1) HR, HRV, AND SpO2

The HR, HRV (RMSSD) and SpO2 calculation as described in Section IV-B6 do not require model training.

#### 2) BP

Because no public dataset was available for BP containing face videos, we decided to train our model using a publicly available PPG-BP dataset released by Liang et al. [56]. The BVP signal is extracted from the PPG signal to train our BP model using the ground truth HR, SBP and DBP data provided in the dataset. Next, we replace the input part of the model architecture to extract the BVP signal from the face video in the same way as we did for the HR, HRV, and SpO₂, and use the remaining part of the pre-trained model to estimate BP.

**TABLE 4.** Performance on PURE Dataset [8].

|  | Steady | Talking | Slow Translation | Fast Translation | Small Rotation | Medium Rotation | Overall |
|---|---|---|---|---|---|---|---|
| **HR (bpm)** | 1.40 | 6.7 | 2.8 | 3.3 | 2.9 | 6.6 | 3.95 |
| **HRV (ms)** | 24.57 | 20.84 | 22.14 | 24.38 | 28.69 | 29.54 | 25.03 |
| **SpO₂ (%)** | 1.48 | 1.53 | 1.78 | 2.08 | 1.39 | 1.56 | 1.64 |

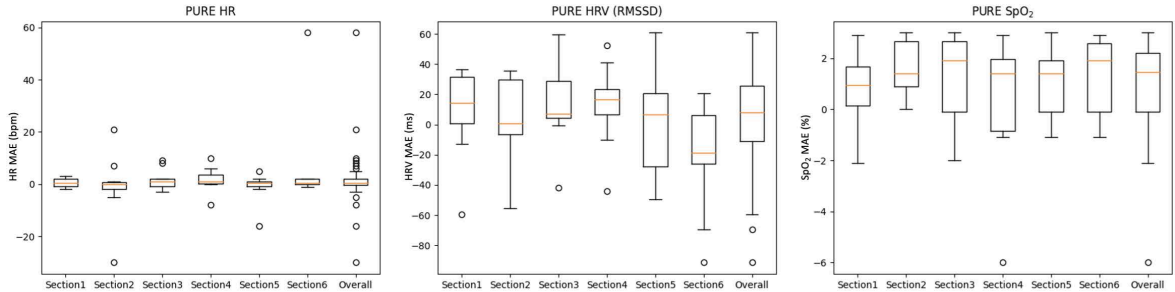

**FIGURE 7.** Boxplot of the MAE in HR, HRV (RMSSD) and SpO₂ calculation on the PURE dataset [8]. The HR estimation is stable overall, though there are some large MAE (over 20bpm). The variation of HRV increases with large motion noise. The SpO₂ maintains between 96%-100% which makes the MAE small.

### 3) DATA PREPROCESSING
First, using the signal skewness metric, the poor signal samples were discarded. Each sample contains 2,100 data points. The sample was clipped to 1,905 data points and segmented into 3 parts. Each part became an input to the network.

### 4) TRAINING AND VALIDATION
Our deep learning model for measuring BP as described in Section IV-B6 was trained on PPG-BP dataset with a batch size of 256 for 50 epochs using Adam optimizer and a learning rate of 0.001. Next we fine-tuned the pre-trained model using the self-created Video-BP Dataset. After a train test split of 90:10, the model was trained with a batch size of 25. The model was trained for 50 epochs with early stopping and using Adam optimizer and a learning rate of 0.0001. K-fold cross validation (k=10) was used during training so that the model gets to train on each type of sample.

## VI. RESULTS
The results of the vital signs calculation are evaluated with MAE. TokyoTech rPPG dataset is applied to evaluate the HR and HRV, and the evaluation results are shown in Table 3. The boxplot of each section is shown in Fig. 6. Since this dataset is collected in a laboratory environment with less light and motion noise, our model achieves 1.73bpm MAE in HR estimation and 18.55ms MAE in HRV estimation. The evaluation results and the boxplot of the PURE dataset are displayed in Table 4 and Fig. 7. Our model achieves better results in steady or small movements such as slow translation and small rotation with less than 3bpm MAE in HR estimation. The MAEs of HRV and SpO₂ calculation are 25.03ms and 1.64% respectively. Our use case Video-HR dataset is built to evaluate the HR measurement from the smartphone. The MAE is 2.49bpm, and the boxplot is shown in Fig. 8. Even though this data has more variation than the laboratory datasets, for less motion noise scenario a good performance was achieved in HR calculations. The BP estimation model achieved an MAE of 6.7mmHg for SBP and 9.6mmHg for DBP on the test set of the Video-BP dataset and the boxplot is shown in Fig. 9.

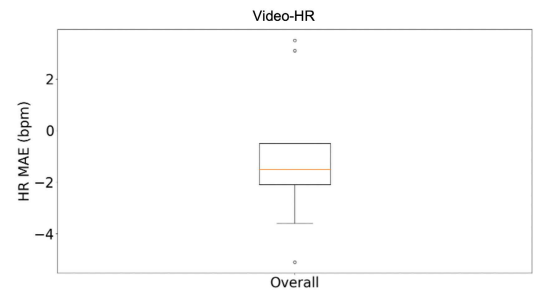

**FIGURE 8.** Boxplot of the MAE in HR calculation on our Video-HR dataset.

### A. DISCUSSION
Our model performs well in HR prediction on all datasets, and MAE of the videos with less motion noise is less than 5bpm, while talking and head movements affect the model stability in some videos. The HRV estimation model depends on the inter peak distance in the BVP signal. The TokyoTech dataset has videos that were recorded under strict controlled laboratory conditions with no movement. Therefore, the MAE for HRV estimation on this dataset is low which is good. However, with increase in the noise in the rPPG data the MAE rises as can be seen in the results of the PURE dataset. In the presence of noise which can be extremely challenging to eliminate, the signal is dampened and its periodicity is affected. Since $SpO_2$ values in the dataset ranged from 90-100%, the model requires more evaluation on special cases such as $SpO_2$ below 90%.

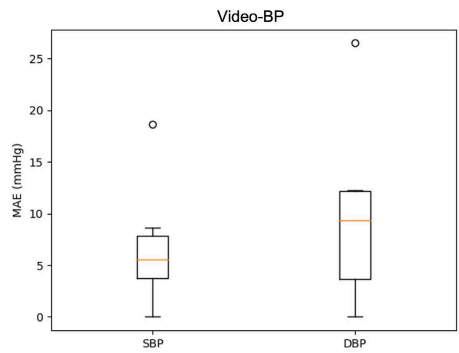

**FIGURE 9.** Boxplot of the MAE in BP calculation on our Video-BP dataset.

**TABLE 5.** Comparing BP estimation model with previous work using Video-BP dataset.

| Model | SBP (mmHg) | DBP (mmHg) |
|---|---|---|
| PKR [58] | 16.4 | 13.3 |
| NN [51] | 13.6 | 12.1 |
| CNN [59] | 14.5 | 10.1 |
| Ours | **6.7** | **9.6** |

Our ResNet BP estimation model returned an MAE of 6.7mmHg and 9.6mmHg for SBP and DBP respectively. This is lower than the MAE of 14.1mmHg and 11.2mmHg for SBP and DBP respectively as reported by Schrumpf et al [52]. However, we tested our models on different datasets. Although the authors [52] stated that they did not find much improvement by using signal derivatives, it reduced the MAE for our model by 1.6mmHg. Since the improvement is not that high, to reduce computational complexity and optimize the use of resources, the model can be trained with only the signal instead of its derivative. The model accuracy may be further improved by training on a larger Video-BP dataset V-D.

## VII. CONCLUSION

In this paper, we extend our previous work [35] and propose an end-to-end framework, ReViSe, to measure users' vital signs (namely HR, HRV, $SpO_2$ and BP) by using a smartphone camera. Our video processing and vital signs estimation methodology are evaluated on two public rPPG datasets. For HR estimation, the model achieved an MAE of 1.48bpm and 4.20bpm with the two public TokyoTech and PURE datasets respectively. We also evaluated the end-to-end framework with our own Video-HR dataset and achieved an MAE of 2.49bpm. HRV heavily relies on a clean BVP signal which is very sensitive to variation in illumination and motion noise. The model achieved an MAE of 1.60% for $SpO_2$ on the PURE dataset. The BP model was trained using a publicly available PPG-BP dataset and tested using a self-created Video-BP dataset. It achieved an MAE of 6.7mmHg and 9.6mmHg for SBP and DBP respectively on the Video-BP dataset.

Medical professionals can use our smartphone applications to measure a patient's vital signs remotely for medical advising. It is a good alternative to invasive methods that are currently the gold standard for vital signs measurement. During the COVID-19 pandemic when in-person medical visits were restricted, such remote contactless methods proved to be extremely useful. In future, the framework can be extended for health screening, healthy risk prediction, or remote heath care provisioning and thereby, making healthcare more affordable and accessible online around the clock.

### 1) FUTURE WORK

Compared to our previous work [35], we propose a more robust noise reduction pipeline, and eliminate the motion noise by detecting user movement and displaying a notification to stay still. Involuntary facial movements are compensated for by using signal processing filters. However, there are many challenges in processing daily living video data. In the future, more research is needed to eliminate noise from facial movements through filtering. This will be helpful in measuring the vital signs of patients who cannot remain still due to neurological disorders. It could also be used for remotely monitoring patient's physiological condition during psychiatric sessions.

Lighting conditions offer difficult challenges in applying the technology in daily living environment. Our model has been evaluated in daily living environment with different scenarios and light conditions. More experiments may be conducted to evaluate the influence of the light variance. A good quality video with higher resolution can provide better accuracy but would take longer time to transmit to the server, process and greater network bandwidth. In collaboration with our industry partner, we had the opportunity to explore and address some of these challenges in this preliminary version of the work. In the future work, we intend to address the above challenges to further reduce the noise, improve the measurement accuracy, test the framework for a wide range of participants of varying age, skin color, gender and health conditions at a clinical environment to collect a greater range of measured vital signs. We plan to use Health Canada approved medical devices to collect the ground truth data for better training and validation of the models. With the larger dataset, we plan to implement deep learning-based models for estimating all the vital signs. We will also extend the framework to support multiple devices at the front-end and perform a usability study of the framework.

## ACKNOWLEDGMENT
The work was done when Amtul Haq Ayesha was at Queen's University.

*(Donghao Qiao and Amtul Haq Ayesha contributed equally to this work.)*

## REFERENCES

[1] C. Wang, T. Pun, and G. Chanel, "A comparative survey of methods for remote heart rate detection from frontal face videos," *Frontiers Bioeng. Biotechnol.*, vol. 6, p. 33, May 2018.

[2] P. V. Rouast, M. T. P. Adam, R. Chiong, D. Cornforth, and E. Lux, "Remote heart rate measurement using low-cost RGB face video: A technical literature review," *Frontiers Comput. Sci.*, vol. 12, no. 5, pp. 858–872, 2016.

[3] F. Shaffer and J. P. Ginsberg, "An overview of heart rate variability metrics and norms," *Frontiers Public Health*, vol. 5, p. 258, Sep. 2017.

[4] T. Tamura, "Current progress of photoplethysmography and SPO2 for health monitoring," *Biomed. Eng. Lett.*, vol. 9, no. 1, pp. 21–36, Feb. 2019.

[5] T. Tamura, Y. Maeda, M. Sekine, and M. Yoshida, "Wearable photoplethysmographic sensors—Past and present," *Electronics*, vol. 3, no. 2, pp. 282–302, 2014.

[6] Y. Sun and N. Thakor, "Photoplethysmography revisited: From contact to noncontact, from point to imaging," *IEEE Trans. Biomed. Eng.*, vol. 63, no. 3, pp. 463–477, Mar. 2015.

[7] Y. Maki, Y. Monno, K. Yoshizaki, M. Tanaka, and M. Okutomi, "Inter-beat interval estimation from facial video based on reliability of BVP signals," in *Proc. 41st Annu. Int. Conf. IEEE Eng. Med. Biol. Soc. (EMBC)*, Jul. 2019, pp. 6525–6528.

[8] R. Stricker, S. M'uller, and H.-M. Gross, "Non-contact video-based pulse rate measurement on a mobile service robot," in *Proc. 23rd IEEE Int. Symp. Robot Hum. Interact. Commun.*, Aug. 2014, pp. 1056–1062.

[9] X. Li, J. Chen, G. Zhao, and M. Pietikainen, "Remote heart rate measurement from face videos under realistic situations," in *Proc. IEEE Conf. Comput. Vis. Pattern Recognit.*, Jun. 2014, pp. 4264–4271.

[10] M.-Z. Poh, D. J. McDuff, and R. W. Picard, "Non-contact, automated cardiac pulse measurements using video imaging and blind source separation," *Opt. Exp.*, vol. 18, no. 10, pp. 10762–10774, 2010.

[11] M.-Z. Poh, D. J. McDuff, and R. W. Picard, "Advancements in noncontact, multiparameter physiological measurements using a webcam," *IEEE Trans. Biomed. Eng.*, vol. 58, no. 1, pp. 7–11, Jan. 2010.

[12] S. Kwon, H. Kim, and K. S. Park, "Validation of heart rate extraction using video imaging on a built-in camera system of a smartphone," in *Proc. Annu. Int. Conf. IEEE Eng. Med. Biol. Soc.*, Aug. 2012, pp. 2174–2177.

[13] A. Gudi, M. Bittner, R. Lochmans, and J. van Gemert, "Efficient real-time camera based estimation of heart rate and its variability," in *Proc. IEEE Int. Conf. Comput. Vis. Workshops*, Oct. 2019, pp. 1570–1579.

[14] A. Gudi, M. Bittner, and J. V. van Gemert, "Real-time webcam heart-rate and variability estimation with clean ground truth for evaluation," *Appl. Sci.*, vol. 10, no. 23, p. 8630, 2020.

[15] M. Kumar, A. Veeraraghavan, and A. Sabharwal, "DistancePPG: Robust non-contact vital signs monitoring using a camera," *Biomed. Opt. Exp.*, vol. 6, no. 5, pp. 1565–1588, 2015.

[16] Z. Wang, X. Yang, and K. T. Cheng, "Accurate face alignment and adaptive patch selection for heart rate estimation from videos under realistic scenarios," *PloS One*, vol. 13, no. 5, 2018, Art. no. e0197275.

[17] S. Tulyakov, X. Alameda-Pineda, E. Ricci, L. Yin, J. F. Cohn, and N. Sebe, "Self-adaptive matrix completion for heart rate estimation from face videos under realistic conditions," in *Proc. IEEE Conf. Comput. Vis. Pattern Recognit.*, Jun. 2016, pp. 2396–2404.

[18] M. E. Tipping and C. M. Bishop, "Probabilistic principal component analysis," *J. Roy. Statist. Soc. B*, vol. 61, no. 3, pp. 611–622, 1999.

[19] A. Hyvärinen and E. Oja, "Independent component analysis: Algorithms and applications," *Neural Netw.*, vol. 13, nos. 4–5, pp. 411–430, 2000.

[20] M. P. Tarvainen, P. O. Ranta-Aho, and P. A. Karjalainen, "An advanced detrending method with application to HRV analysis," *IEEE Trans. Biomed. Eng.*, vol. 49, no. 2, pp. 172–175, Feb. 2002.

[21] P. D. Welch, "The use of fast Fourier transform for the estimation of power spectra: A method based on time averaging over short, modified periodograms," *IEEE Trans. Audio Electroacoust.*, vol. AE-15, no. 2, pp. 70–73, Jun. 1967.

[22] V. Bazarevsky, Y. Kartynnik, A. Vakunov, K. Raveendran, and M. Grundmann, "BlazeFace: Sub-millisecond neural face detection on mobile GPUs," 2019, *arXiv:1907.05047*.

[23] I. Grishchenko, A. Ablavatski, Y. Kartynnik, K. Raveendran, and M. Grundmann, "Attention mesh: High-fidelity face mesh prediction in real-time," 2020, *arXiv:2006.10962*.

[24] J. Przybyło, "A deep learning approach for remote heart rate estimation," *Biomed. Signal Process. Control*, vol. 74, Apr. 2022, Art. no. 103457.

[25] R. Song, H. Chen, J. Cheng, C. Li, Y. Liu, and X. Chen, "PulseGAN: Learning to generate realistic pulse waveforms in remote photoplethysmography," *IEEE J. Biomed. Health Informat.*, vol. 25, no. 5, pp. 1373–1384, May 2021.

[26] Y. Y. Tsou, Y. A. Lee, C. T. Hsu, and S. H. Chang, "Siamese-rPPG network: Remote photoplethysmography signal estimation from face videos," in *Proc. 35th Annu. ACM Symp. Appl. Comput.*, Mar. 2020, pp. 2066–2073.

[27] M. B. Garcia, N. U. Pilueta, and M. F. Jardiniano, "VITAL APP: Development and user acceptability of an IoT-based patient monitoring device for synchronous measurements of vital signs," in *Proc. IEEE 11th Int. Conf. Humanoid, Nanotechnol., Inf. Technol., Commun. Control, Environ., Manag. (HNICEM)*, Nov. 2019, pp. 1–6.

[28] G. Balakrishnan, F. Durand, and J. Guttag, "Detecting pulse from head motions in video," in *Proc. IEEE Conf. Comput. Vis. Pattern Recognit.*, Jun. 2013, pp. 3430–3437.

[29] R. Irani, K. Nasrollahi, and T. B. Moeslund, "Improved pulse detection from head motions using DCT," in *Proc. Int. Conf. Comput. Vis. Theory Appl. (VISAPP)*, vol. 3, Jan. 2014, pp. 118–124.

[30] H. Monkaresi, R. A. Calvo, and H. Yan, "A machine learning approach to improve contactless heart rate monitoring using a webcam," *IEEE J. Biomed. Health Informat.*, vol. 18, no. 4, pp. 1153–1160, Jul. 2014, doi: 10.1109/JBHI.2013.2291900.

[31] A. Osman, J. Turcot, and R. El Kaliouby, "Supervised learning approach to remote heart rate estimation from facial videos," in *Proc. 11th IEEE Int. Conf. Workshops Autom. Face Gesture Recognit. (FG)*, vol. 1, May 2015, pp. 1–6.

[32] W. Chen and D. McDuff, "Deepphys: Video-based physiological measurement using convolutional attention networks," in *Proc. Eur. Conf. Comput. Vis. (ECCV)*, Sep. 2018, pp. 349–365.

[33] R. Špetlík, V. Franc, and J. Matas, "Visual heart rate estimation with convolutional neural network," in *Proc. Brit. Mach. Vis. Conf.*, Newcastle, U.K., Sep. 2018, pp. 3–6.

[34] A. K. Kanva, C. J. Sharma, and S. Deb, "Determination of SpO2 and heart-rate using smartphone camera," in *Proc. Int. Conf. Control, Instrum., Energy Commun. (CIEC)*, Jan. 2014, pp. 237–241.

[35] D. Qiao, F. Zulkernine, R. Masroor, R. Rasool, and N. Jaffar, "Measuring heart rate and heart rate variability with smartphone camera," in *Proc. 22nd IEEE Int. Conf. Mobile Data Manage. (MDM)*, Jun. 2021, pp. 248–249.

[36] A. H. Ayesha, "ReViSe: An end-to-end framework for remote measurement of vital signs," M.S. dissertation, Queen's Univ., Kingston, ON, Canada, 2022.

[37] C. El-Hajj and P. A. Kyriacou, "A review of machine learning techniques in photoplethysmography for the non-invasive cuff-less measurement of blood pressure," *Biomed. Signal Process. Control*, vol. 58, Apr. 2020, Art. no. 101870.

[38] L. Xi, W. Chen, C. Zhao, X. Wu, and J. Wang, "Image enhancement for remote photoplethysmography in a low-light environment," in *Proc. 15th IEEE Int. Conf. Autom. Face Gesture Recognit. (FG)*, Nov. 2020, pp. 1–7.

[39] A. Nemcovaa, I. Jordanovaa, M. Vareckaa, R. Smiseka, L. Marsanovaa, L. Smitala, and M. Viteka, "Monitoring of heart rate, blood oxygen saturation, and blood pressure using a smartphone," *Biomed. Signal Process. Control*, vol. 59, May 2020, Art. no. 101928.

[40] M. Chowdhury, M. Shuzan, M. Chowdhury, Z. Mahbub, M. M. Uddin, A. Khandakar, and M. I. Reaz, "Estimating blood pressure from the photoplethysmogram signal and demographic features using machine learning techniques," *Sensors*, vol. 20, no. 11, p. 3127, 2020.

[41] K. Kira and L. A. Rendell, "The feature selection problem: Traditional methods and a new algorithm," in *Proc. AAAI*, vol. 2, Jul. 1992, pp. 129–134.

[42] X. Xing and M. Sun, "Optical blood pressure estimation with photoplethysmography and FFT-based neural networks," *Biomed. Opt. Exp.*, vol. 7, pp. 3007–3020, Aug. 2016.

[43] P.-W. Huang, C.-H. Lin, M.-L. Chung, T.-M. Lin, and B.-F. Wu, "Image based contactless blood pressure assessment using pulse transit time," in *Proc. Int. Autom. Control Conf. (CACS)*, Nov. 2017, pp. 1–6, doi: 10.1109/CACS.2017.8284275.

[44] H. Luo, D. Yang, A. Barszczyk, N. Vempala, J. Wei, S. Wu, P. Zheng, G. Fu, K. Lee, and Z. Feng, "Smartphone-based blood pressure measurement using transdermal optical imaging technology," *Cardiovascular Imag.*, vol. 12, no. 8, 2019, Art. no. e008857.

[45] H. Rahman, M. Ahmed, S. Begum, and P. Funk, "Real time heart rate monitoring from facial RGB color video using webcam," in *Proc. 9th Annu. Workshop Swedish Artif. Intell. Soc. (SAIS)*, Jun. 2016, pp. 15–46.

[46] W. Verkruysse, L. Svaasand, and J. Nelson, "Remote plethysmographic imaging using ambient light," *Opt. Exp.*, vol. 16, no. 26, pp. 21434–21445, 2008.

[47] M. Lewandowska, J. Ruminski, T. Kocejko, and J. Nowak, "Measuring pulse rate with a webcam—A non-contact method for evaluating cardiac activity," in *Proc. Federated Conf. Comput. Sci. Inf. Syst. (FedCSIS)*, Sep. 2011, pp. 405–410.

[48] P. Su, X.-R. Ding, Y.-T. Zhang, J. Liu, F. Miao, and N. Zhao, "Long-term blood pressure prediction with deep recurrent neural networks," in *Proc. IEEE EMBS Int. Conf. Biomed. Health Informat. (BHI)*, Mar. 2018, pp. 323–328.

[49] G. Slapnicar, N. Mlakar, and M. Lustrek, "Blood pressure estimation from photoplethysmogram using a spectro–temporal deep neural network," *Sensors*, vol. 19, no. 15, p. 3420, 2019.

[50] S. Shimazaki, S. Bhuiyan, H. Kawanaka, and K. Oguri, "Features extraction for cuffless blood pressure estimation by autoencoder from photoplethysmography," in *Proc. 40th Annu. Int. Conf. IEEE Eng. Med. Biol. Soc. (EMBC)*, Jul. 2018, pp. 2857–2860.

[51] C. G. Viejo, S. Fuentes, D. Torrico, and F. Dunshea, "Non-contact heart rate and blood pressure estimations from video analysis and machine learning modelling applied to food sensory responses: A case study for chocolate," *Sensors*, vol. 18, no. 6, p. 1802, Jun. 2018.

[52] F. Schrumpf, P. Frenzel, C. Aust, G. Osterhoff, and M. Fuchs, "Assessment of non-invasive blood pressure prediction from PPG and RPPG signals using deep learning," *Sensors*, vol. 21, no. 18, p. 6022, 2021.

[53] X. Guo, Y. Li, and H. Ling, "Lime: Low-light image enhancement via illumination map estimation," *IEEE Trans. Image Process.*, vol. 26, no. 2, pp. 982–993, Feb. 2017.

[54] G. Quellec, M. Lamard, P. H. Conze, P. Massin, and B. Cochener, "Automatic detection of rare pathologies in fundus photographs using few-shot learning," *Med. Image Anal.*, vol. 61, Apr. 2020, Art. no. 101660.

[55] G. Haan and V. Jeanne, "Robust pulse-rate from chrominance-based rPPG," *IEEE Trans. Biomed. Eng.*, vol. 60, no. 10, pp. 2878–2886, Oct. 2013.

[56] M. Elgendi, V. Galli, C. Ahmadizadeh, and C. Menon, "Dataset of psychological scales and physiological signals collected for anxiety assessment using a portable device," *Data*, vol. 7, no. 9, p. 132, 2022.

[57] X. Fan, Q. Ye, X. Yang, and S. Choudhury, "Robust blood pressure estimation using an RGB camera," *J. Ambient Intell. Humanized Comput.*, vol. 11, no. 11, pp. 4329–4336, 2018.

[58] M. Jain, S. Deb, and A. V Subramanyam, "Face video based touchless blood pressure and heart rate estimation," in *Proc. IEEE 18th Int. Workshop Multimedia Signal Process. (MMSP)*, Sep. 2016, pp. 1–5.

[59] E. Brophy, W. Muehlhausen, A. F. Smeaton, and T. E. Ward, "Optimised convolutional neural networks for heart rate estimation and human activity recognition in wrist Worn sensing applications," 2020, *arXiv:2004.00505*.

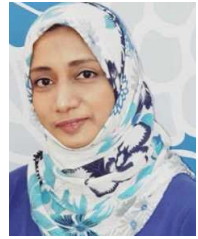

**AMTUL HAQ AYESHA** received the M.Sc. degree in the field of study in artificial intelligence from the School of Computing, Queen's University. She has worked on the Veyetals application with Markitech as a Research Intern. She is currently a Software Engineer with Sanofi Pasteur. She has won the 2022 3MT at Queen's University and represented the university at the Ontario 3MT. She aspires to pursue her Doctorate studies in using machine learning techniques in health care.

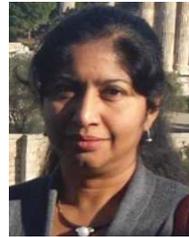

**FARHANA ZULKERNINE** (Member, IEEE) is currently an Associate Professor, the Coordinator of the Cognitive Science Program, and the Director of the Big Data Analytics and Management Laboratory, School of Computing, Queen's University. She is also a Certified Professional Engineer and has worked in three continents in research and development collaborating with many industry and academic partners to develop usable technology. She serves on multiple funding and conference committees and has published over a 100 research articles. Her research interests include artificial intelligence and cognitive computing modeling for big data analytics and management.

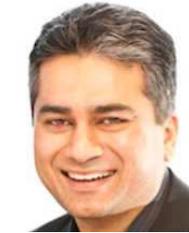

**NAUMAN JAFFAR** is currently a Passionate and Serial Entrepreneur with a background in technology marketing and digital transformation. He has established more than six healthcare, telco, consulting, and SaaS firms in the previous years and has overseen more than 50 healthcare AI and IT projects in his entrepreneurial and corporate experience. He is also the Founder of Your Doctors Online, MarkiTech.AI, SenSights.AI, Veyetals.com, and a number of innovative niche solutions in health and wellness utilizing powerful new technologies, such as AI, machine learning, and IoT. The majority of the solutions and platforms are aimed at seniors, older adults, and community care. He wants to sustain the company's success by making a difference in Age Tech, helping seniors age in place, achieving hospital at home trends, and helping over 25K companies in North America overcome digital transformation challenges with some very innovative, affordable, and simple-to-use solutions.

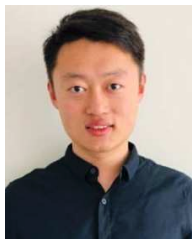

**DONGHAO QIAO** is currently pursuing the Ph.D. degree with the School of Computing, Queen's University, Canada. His Ph.D. research focuses on medical data processing and traffic data processing with artificial intelligence. His research interests include machine learning, deep learning, computer vision, natural language processing, and data mining.

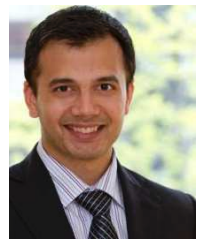

**RAIHAN MASROOR** is currently a Techie who is always looking for newer, better, and faster ways to solve problems and improve people's lives. He worked at BlackBerry designing/creating social applications, then evolved in developing new services for businesses at TELUS. With his passion and expertise in technology and digital marketing, coupled with his mission to make people's lives better, he is also leading Your Doctors Online to bring affordable healthcare to all corners of the globe.

• • •